# Extending Term Subsumption Systems for Uncertainty Management*

John Yen

Department of Computer Science
Texas A&M University
College Station, TX 77843

Piero P. Bonissone

Artificial Intelligence Program
General Electric Corporate Research and Development
P.O. Box 8
Schenectady, NY 12308

## Abstract

A major difficulty in developing and maintaining very large knowledge bases originates from the variety of forms in which knowledge is made available to the KB builder. The objective of this research is to bring together two complementary knowledge representation schemes: *term subsumption* languages, which represent and reason about defining characteristics of concepts, and *approximate reasoning* models, which deal with uncertain knowledge and data in expert systems. Previous works in this area have primarily focused on probabilistic inheritance. In this paper, we address two other important issues regarding the integration of term subsumption-based systems and approximate reasoning models. First, we outline a general architecture that specifies the interactions between the deductive reasoner of a term subsumption system and an approximate reasoner. Second, we generalize the semantics of terminological language so that terminological knowledge can be used to make plausible inferences. The architecture, combined with the generalized semantics, forms the foundation of a synergistic tight integration of term subsumption systems and approximate reasoning models.

## 1 Introduction

Current expert system technology does not provide enough support for the development and maintenance of very large knowledge bases. One of the major impediments for achieving this task is the variety of forms in which the available knowledge is expressed [AB87]. Squeezing all the knowledge required by an expert system into one or two representation formalisms is difficult, time-consuming, and usually an inadequate solution to the task at hand. Hence, there is a need to integrate multiple knowledge representation schemes. The objective of this research is to bring together two such schemes: *term subsumption* languages, which represent and reason about defining characteristics of concepts, and *approximate reasoning* models, which deal with uncertain knowledge and data in expert systems. The management of uncertainty is an important issue in expert systems because contents of a knowledge base are often incomplete, imprecise, and uncertain. For example, most medical expert system uses judgemental knowledge (i.e., rules of thumb) of physicians, which is known to be uncertain, for diagnosing a patient's disease.

*Term Subsumption Languages* refers to knowledge representation formalisms that employ a formal language, with a formal semantics, for the definition of terms (more commonly referred to as concept or classes), and that deduce whether one term subsumes (is more general than) another [POK90]. These formalisms generally descend from the ideas presented in KL-ONE [BS85]. Term subsumption languages are a generalization of both semantic networks and frames because the languages have well-defined semantics, which is often missing from frames and semantic networks [Woo75, Bra83].

The major strength of term subsumption systems is their reasoning capabilities offered by a *classifier*. The classifier is a special purpose reasoner that automatically infers and maintains a consistent and accurate taxonomic lattice of logical subsumption relations between concepts [SL83]. Even though deductive inference performed by the classifier can be viewed as a special kind of approximate reasoning, it is more efficient to use the classifier rather than triggering a long chain of rules.

Previous works in the area of uncertainty management in frame-based systems have been focusing on the inheritance of probabilistic knowledge in a concept taxonomy [HO88, Gro86]:

> Given that A is subsumed by B, B implies C with certain degree of belief, and x is an instance

*This work was partially supported by the Defense Advanced Research Projects Agency (DARPA) under USAF/Rome Air Development Center contract F30602-85-C-0033. Views and conclusions contained in this paper are those of the authors and should not be interpreted as representing the official opinion or policy of DARPA or the U.S. Government.

of A to certain degree. What can we say about the likelihood that x is an instance of C?

Two other important issues, however, remain to be addressed for a complete integration of term subsumption-based systems and approximate reasoning models. First, we need to develop a general architecture that specifies the interactions between two reasoning components of the system: a deductive reasoner and an approximate reasoner. Second, we need to specify how to use definitional (terminological) knowledge for approximate reasoning. This paper outlines our approach to address these issues.

## 2 Related Work

Lokendra Shastri has developed a framework, based on the principle of maximum entropy, for dealing with uncertainty in semantic networks [Sha85, Sha89]. His approach is based on the assumption that the system has certain statistical data (e.g., the number of red apples, the number of sweet apples, ...). Based on these statistical data, Shastri's evidential theory answers questions of the following kind: *Given that an instance, x, is red and sweet, is x more likely to be an apple or a grape ?* The major problem of applying Shastri's theory to expert systems lie in the lack of statistical data. For example, most medical expert systems are based on knowledge elicited from human experts, not a patient database. Therefore, it is very difficult to obtain those marginal probability judgements used in Shastri's model (e.g., the total number of patients having seropositive rheumatoid arthritis).

A recent work by Heinsohn and Owsnicki-Klewe proposes a model of probabilistic reasoning in hybrid term subsumption systems [HO88]. Uncertain knowledge is represented as *probabilistic implications* in the form of $C_1 \Rightarrow_s C_2$ where s denotes the conditional probability $P(C_2(x)|C_1(x))$, $C_1$ and $C_2$ are concepts defined in the terminological knowledge base. The reasoning mechanism of their model is *probabilistic inheritance* (i.e., the inheritance of probabilistic implications in concept taxonomy). The issue of non-monotonicity of probabilistic inheritance has also been discussed in [Gro86].

Even though Heinsohn and Owsnicki-Klewe's model, as they claim, enlarges the range of applicability of hybrid term subsumption systems, it is limited in the kind of uncertain knowledge it can represent. Most uncertain rules in expert systems consist of complex conditions, which can not all be represented as concept definitions. Therefore, probabilistic implications need to be extended to express complex conditions before they can be applied to expert systems. Our previous work in developing a production systems, CLASP, on top of LOOM has extended LOOM's pattern matcher for handling complex conjunctive conditions [YNM89]. This provides us the basis for incorporating PRIMO's plausible rules into LOOM.

## 3 Extending Term Subsumption Systems for Approximate Reasoning

In this section, we outline a general architecture for incorporating approximate reasoning into term subsumption systems. The architecture is based on the following extended assertional language that allow the assertion and retraction of uncertain statements:

```
<assertion> ::= (tell (<literal> x))
              | (tell <literal>)
              | (forget <literal> )
<query> ::= (ask <literal>)
<literal> ::= (<role> <instance> <instance>)
            | (<concept> <instance>)
```

where x is a measure of certainty degree. Depending on the approximate reasoning model used, x can represent the degree of membership in a fuzzy set, a probability, an interval probability, or a basic probability value in the Dempster-Shafer framework. Using such a generic assertional language enables us to describe the interaction between major components of the architecture in a way that is independent of the choice of approximate reasoning models (e.g., Bayesian probabilistic reasoning, fuzzy reasoning, and Dempster-Shafer reasoning).

The architecture specifies the function and the interaction between three major components: a deductive reasoner, an approximate reasoner, an assertion analyzer, and a query analyzer. Both the deductive reasoner and the approximate reasoner performs inference and maintain the consistency of the facts database. The deductive reasoner performs logic deductions based on terminological knowledge and certain facts. The approximate reasoner performs plausible inference based on plausible rules, uncertain facts, and terminological knowledge.

The interaction between the two reasoners is the following: The deductive reasoner informs the approximate reasoner about all changes made to the facts database. The approximate reasoner, on the other hand, informs deductive reasoner only about the addition or removal of facts whose certainty degrees are ones.

The assertion analyzer determines how to translate a user's assertional statements into assertional changes (i.e., tell and forget statements) to the deductive reasoner and the approximate reasoner. For example, suppose that the system is first informed that John is an instance of a concept rich-person. But the user later informs the system that John is rich only with 0.8 certainty degree. The assertion analyzer will translate the second assertion (`tell ((Rich-person John) 0.8)`) into two internal assertional changes: a retraction (`forget (Rich-person John)` ) to the deductive reasoner and an assertion `(tell ((Rich-person John) 0.8))` to the approximate reasoner.

The query processor generates responses to external queries by retrieving facts in the database or by invoking the approximate reasoner and the deductive reasoner to perform goal-driven inference. In order to construct appropriate answers using facts generated by multiple sources (i.e., the user, the deductive reasoner, and the approximate reasoner), the query processor needs to distinguish three types of facts in the database:

- *asserted* facts
- facts *deduced* from the deductive reasoner
- facts *inferred* by the approximate reasoner.

Asserted facts take precedence over facts deduced by the deductive reasoner, which take precedence over facts inferred by the approximate reasoner. This can be illustrated using the following example:

Plausible Rule:
    If a person drives a Mercedes,
    then he/she is likely (0.8) to be rich.

Terminological Knowledge:
    A person is rich if and only if
    he/she lives in a mansion.

```
(tell (Drives John car-1))
(tell (Mercedes car-1))
(ask (Rich John))
     Answer: John is likely (0.8) to be rich.
(tell (Live-in John house-1))
(tell (Mansion house-1))
(ask (Rich John))
     Answer: John is rich.
(forget (Live-in John house-1))
(ask (Rich John))
     Answer: John is likely (0.8) to be rich.
```

Therefore, when the fact that John is rich is deduced by the deductive reasoner, it overrides the plausible conclusion that John is likely (0.8) to be rich. But when the deduced fact is retracted, the plausible conclusion is used again by the query processor as the answer.

## 3.1 PRIMO and LOOM

Another experiment in extending a term-subsumption system for approximate reasoning is currently under way. The LOOM system [Mac88] is being loosely integrated with the Plausible ReasonIng MOdule (PRIMO) [BCG89], [ABS90]. In this experiment, LOOM maintains precise concept definitions as taxonomical knowledge in the TBox. Fully grounded instances of these concepts are maintained in LOOM's assertional component (ABox) and *shadowed*, i.e., duplicated and independently maintained, as nodes in a PRIMO graph. This partition allows us to first exploit the definitional knowledge in the Tbox, obtaining all the *strong* deductions derivable from such taxonomical knowledge. Default rules, based on concept specificity, can also be efficiently executed by using LOOM's subsumption test mechanism. After these deductions have been completed in the TBox (at the first order predicate calculus level), they are inherited by the concepts *instances*. This information is passed to PRIMO, which maintains the truth state of each instance as a node in a PRIMO graph, to establish the context and the input from which the approximate inferences will be made. In this scheme, PRIMO will then execute the applicable plausible and default rules and augment the value assignment of each propositional variable with the derived values. In the rest of this section we will briefly summarize the approximate reasoning capabilities that PRIMO can provide in this architecture.

**PRIMO** PRIMO is a software tool that integrates the theories of defeasible reasoning (based on default values supported by nonmonotonic rules) with plausible reasoning (based on monotonic rules with degrees of uncertainty). PRIMO has a reasoning system (a language for representing uncertain and default knowledge, along with algorithms for reasoning in this language) and a computing environment.

**Possibilistic Reasoning in PRIMO** The uncertainty representation used in PRIMO is based on the semantics of many-valued logics. PRIMO, like its predecessor RUM [BGD87], uses a combination of fuzzy logic and interval logic to represent and reason about uncertainty. This approach has been successfully demonstrated in two DARPA applications, the Pilot's Associate and Submarine Operational Automation System programs.

PRIMO handles uncertain information by qualifying each possible value assignment to any given propositional variable with an uncertainty interval. The interval's lower bound represents the minimal degree of confirmation for the value assignment. The upper bound represents the degree to which the evidence failed to refute the value assignment. The interval's width represents the amount of ignorance attached to the value assignment. The uncertainty intervals are propagated and aggregated by Triangular-norm-based uncertainty calculi (see [Bon87, SS63]). The uncertainty interval constrains intervals of subsequent, dependent values. These uncertainty calculi are further elaborated in [Bon89].

**Probabilistic Reasoning in PRIMO** PRIMO can also *emulate* the propagation of probability values throughout a directed acyclic graph. This probabilistic model follows the concept of support logic theory [Bal87], in which imprecise probability values (represented by intervals) can be assigned to input nodes and links in the graph. In this scheme an input node $A$ can be assigned the probability $P(A) = [x_1, x_2]$. Two conditional probabilities can be attached to each link of the graph: $P(B \mid A) = [w_1, w_2]$ and $P(B \mid \neg A) = [w_3, w_4]$. The formula of total probability is used to compute the interval value of the conclusion node $B$. This emulation mode in PRIMO can be used in lieu of the possibilistic reasoning mode described in the previous paragraph.

**Defeasible Reasoning in PRIMO** PRIMO handles incomplete information by evaluating non-monotonic justified (NMJ) rules. These rules are used to express the knowledge engineer's preference in cases of total or partial ignorance regarding the value assignment of a given propositional variable. The NMJ rules are used when there is no plausible evidence (to a given numerical threshold of belief or certainty) to infer that a given value assignment is either true or false. The conclusions of NMJ rules can be retracted by the belief revision system, when enough plausible evidence is available.

PRIMO uses the numerical certainty values generated by plausible reasoning techniques to quantitatively distinguish the admissible extensions generated by defeasible reasoning techniques. The method selects a *maximally consistent extension* (see [BCG89]) given all currently available information.

For efficiency considerations some restrictions are placed on the language in which one can express PRIMO rules. The monotonic rules are non-cyclic Horn clauses, and are maintained by a linear belief revision algorithm operating on a rule graph. The NMJ rules can have cycles, but cannot have disjunctions in their conclusions.

By identifying sets of NMJ rules as strongly connected components (SCC's), we can decompose the rule graph into a directed acyclic graph (DAG) of nodes, some of which are SCCs with several input edges and output edges. PRIMO contains algorithms to efficiently propagate uncertain and incomplete information through these structures at run time. Treating the SCCs independently can result in a significant performance improvement over processing the entire graph. However, this heuristic may result in loss of correctness in the worst case. These algorithms require finding satisfying assignments for nodes in each SCC, and are thus NP-hard in the unrestricted case. We can achieve tractability by restricting the size and complexity of the SCC's, precomputing their structural information, and using run-time evaluated certainty measures to select the most likely extension.

## 4 Using Terminological Knowledge for Approximate Reasoning

We will now focus our discussion on the representation and propagation of uncertainty in the terminological knowledge. We will use the following example to illustrate how definitional knowledge, which is expressed in term subsumption languages, can be used for making plausible inference. Suppose **Successful-father** is defined as a father all whose children are college graduates. This can be expressed as[1]

```
(defconcept Father
    (:and Male (:at-least 1 Child )))
(defconcept Successful-Father (:and Father
    (:all Child College-Graduate)))
```

with the following logic interpretation:

$\forall x$ Successful-father$(x) \iff$ Father$(x) \wedge$
$\quad [\, \forall y$ Child$(x,y) \Rightarrow$ College-Graduate$(y)\,]$

Based on the inferential power of their classifiers, term subsumption systems tidily handle the pattern matching problem of recognizing John as a successful-father, given facts such as "John is a male person", "John has two children", "Philip is John's son", "Angela is John's daughter", "both Philip and Angela are college graduates". However, if any of the facts are uncertain (e.g., Philip is likely to be a college graduate), the deductive pattern matcher of the term subsumption system will not be able to deduce the likelihood that John is a successful father. To use definitional knowledge to draw this kind of plausible inference, we need to define how to measure the degree to which each constraint in a term-forming expression is satisfied. We will focus our discussion on the term-forming expression shown in Figure 1.

In this paper, we will denote the degree to which an instance x satisfies a terminological expression $\mathcal{E}$ by $\mu_{\mathcal{E}}(x)$. It is obvious that the formula for computing $\mu_{\mathcal{E}}(x)$ must be consistent with the semantics of $\mathcal{E}$'s logical interpretation (shown in Figure 1).

### 4.1 Soft Value Restriction

A value restriction in a terminological language (e.g., `(:all Child College-Graduate)`) constrains all the slot values of an object to be instances of a given class. We can generalize this constraint to an "elastic constraint", or "soft constraint", in two ways. The logic implication in the original semantics

$$\forall y\ \text{Child}(x,y) \to \text{College-Graduate}(y) \qquad (1)$$

can be generalized to a fuzzy implication operator. Thus, the degree to which a value restriction `(:all Child College-Graduate)` is satisfied by an instance x is determined by the degree to which the implication is true for x. This can be formulated as follows:

$$\mu_{(:\text{all C CG})}(x) = \inf_{y_i}\left[\mu_{C(x,y) \to CG(y)}(x, y_i)\right] \qquad (2)$$

---

[1] We use the syntax of LOOM knowledge representation system[Mac88] to define concepts and relations in this paper.

| Expression<br>e | Interpretation<br>e |
|---|---|
| :primitive (*concept*) | a unique primitive concept |
| (:and $C_1$ $C_2$) | $\lambda x.\ C_1(x) C_2(x)$ |
| (:and $R_1$ $R_2$) | $\lambda xy.\ R_1(x,y) R_2(x,y)$ |
| (:at-least 1 $R$) | $\lambda x.\ \exists y.\ R(x,y)$ |
| (:all $R$ $C$) | $\lambda x.\ \forall y.\ R(x,y) \rightarrow C(y)$ |

Figure 1: Semantics of Some Term-Forming Expressions

where C and CG are shorthands of Child and College-Graduate respectively, and $\mu_{C(x,y) \rightarrow CG(y)}(x, y_i)$ can be defined using various fuzzy implication operators [MS89].

An alternative approach to generalizing the semantics of a value restriction is to use the notion of conditional possibility in possibility theory [Zad78, DP88]:

$$\mu_{(:\text{all C CG})}(x)$$
$$= 1 - Poss(\neg CG(y)|C(x,y)) \quad (3)$$
$$= 1 - \frac{\max_y \{\min[1 - \mu_{CG}(y), \mu_C(x,y)]\}}{\max_y \mu_C(x,y)} \quad (4)$$

In essence, this formula computes a measure that a child of x is *necessarily* a college graduate. It is easy to verify that both generalizations of the value restriction above are consistent with the original semantics.

### 4.2 Soft Number Restriction

Zadeh has used sigma-counts to define the cardinality of a fuzzy set in his test-score semantics [Zad81]:

$$\Sigma COUNT(A) = \sum_{i=1}^{n} \mu_A(x_i) \quad (5)$$

where A is a fuzzy set characterized by a membership function $\mu_A$. We can thus generalize the number restriction in terminological languages to a "soft" number restriction using sigma-counts and fuzzy numbers:

$$\mu_{(:\text{at-least n R2})}(x) = \mu_{at-least-n}\left(\sum_y \mu_{R2}(x,y)\right) \quad (6)$$

$$\mu_{(:\text{at-most n R2})}(x) = \mu_{at-most-n}\left(\sum_y \mu_{R2}(x,y)\right) \quad (7)$$

where at-least-n and at-most-n are fuzzy subsets of real numbers characterized by the following membership functions:

$$\mu_{at-least-n}(x) = \begin{cases} 0 & x \leq n \\ x - n + 1 & n - 1 \leq x \leq n \\ 1 & x \geq n \end{cases}$$

$$\mu_{at-most-n}(x) = \begin{cases} 1 & x \leq n \\ x - n & n \leq x \leq n + 1 \\ 0 & x \geq n + 1 \end{cases}$$

Finally, the degree an instance satisfies a conjunction of sub-expressions can be computed using the "min" operator in fuzzy set theory. For instance, the degree to which an instance is a Successful-Father can be obtained as follows:

$$\mu_{\text{Successful-Father}}(x) =$$
$$\min\{\mu_{\text{Father}}(x), \mu_{(:\text{all C CG})}(x)\} \quad (8)$$

It should be noted that other operators could be used to represent the conjunction of the sub-expressions. In particular, by using Triangular Norms we could represent the lower and upper bounds of such intersection as:

$$\mu_{\text{Successful-Father}}(x) =$$
$$[\max\{0, \mu_{\text{Father}}(x) + \mu_{(:\text{all C CG})}(x) - 1\},$$
$$\min\{\mu_{\text{Father}}(x), \mu_{(:\text{all C CG})}(x)\}] \quad (9)$$

By extending the terminological knowledge to represent and propagate uncertain information, we will be able to generate approximate deductions about the instances of the concepts defined in the Tbox. This uncertain information could then be given as input to an approximate reasoner such as PRIMO, as we described in the previous section.

## 5 Summary

We have outlined a general architecture that extends term subsumption systems for uncertainty management. We have described PRIMO, an approximate reasoner that is currently being loosely integrated with LOOM to test such an architecture. We have also generalized the semantics of terminological languages so that they can be used for drawing plausible inferences. The integration of terminological capabilities with approximate reasoning offers several important benefits. First, it facilitates the application of term subsumption systems to expert systems. Second, it enhances the reusability of terminological knowledge because it is used for deductive reasoning as well as approximate reasoning. Third, it improves the maintainability and the explanation capabilities of expert systems because the meanings of terms are explicitly represented and are separated from heuristic knowledge that is used for plausible inferences.

# Session 9:

## Belief Networks: Approximate Methods, Intervals, Simulations